\documentclass[10pt,twocolumn,letterpaper]{article}

\usepackage{iccv}
\usepackage{times}
\usepackage{epsfig}
\usepackage{graphicx}
\usepackage{amsmath}
\usepackage{amssymb}

\usepackage{url}

\usepackage{multirow}

\usepackage{xfrac} 

\newcommand{\figref}[1]{Fig.~\ref{#1}}
\newcommand{\eqreff}[1]{Eq.~\eqref{#1}}
\newcommand{\out}{${\widehat{\text{Y}}}$}
\newcommand{\outf}{\widehat{\text{Y}}}
\newcommand{\refdataset}{\ificcvfinal\cite{datasetDL}\else\cite{datasetDL_anon}\fi}

\usepackage[pagebackref=true,breaklinks=true,letterpaper=true,colorlinks,bookmarks=false]{hyperref}

\iccvfinalcopy 

\def\iccvPaperID{27} 

\ificcvfinal\fi

\begin{document}

\title{Architectural Tricks for Deep Learning in Remote Photoplethysmography}

\author{Mikhail Kopeliovich \hspace{35pt} Yuriy Mironenko \hspace{35pt} Mikhail Petrushan\\
Southern Federal University, Center of Neurotechnologies\\
Rostov-on-Don, Russian Federation\\
{\tt\small kop@km.ru}
}

\maketitle
\ificcvfinal\thispagestyle{empty}\fi

\begin{abstract}
   Architectural improvements are studied for convolutional network performing estimation of heart rate (HR) values on color signal patches. Color signals are time series of color components averaged over facial regions recorded by webcams in two scenarios: Stationary (without motion of a person) and Mixed Motion (different motion patterns of a person). HR estimation problem is addressed as a classification task, where classes correspond to different heart rate values within the admissible range of [40;~125]~bpm. Both adding convolutional filtering layers after fully connected layers and involving combined loss function where first component is a cross entropy and second is a squared error between the network output and smoothed one-hot vector, lead to better performance of HR estimation model in Stationary and Mixed Motion scenarios.
\end{abstract}

\section{Introduction}
Wide-spreading of cheap color cameras, especially built into smartphones,
makes efforts to retrieve biosignals remotely, out of video and without specials sensors,
very appealing.

Early papers~\cite{Verkruysse2008,Wieringa2005,Wu2003} demonstrated this possibility,
with information retrieved trough analysis of small fluctuations of skin color. This approach was later called iPPG (imaging photoplethysmography) or rPPG (remote photoplethysmography),
which are effectively synonyms (possibly, except the specific cases like one-pixel camera~\cite{Wang2019}).
Later papers also demonstrated analysis of micro-movements caused by pulse (imaging ballistocardiography)~\cite{Balakrishnan2013,Shao2017}.

More recent works demonstrated that pulse signal can be recovered
from long distances~\cite{Shi2010} up to 50~m~\cite{Blackford2016} and from images as small as $41\times30$ pixels~\cite{McDuff_2018_CVPR_Workshops}. 

Motion of the subject is a significant challenge.
The standard approach is to extract heart rate~(HR) information from a signal based on small fluctuations of skin color
(the green color component is mostly useful~\cite{Osman2015,Verkruysse2008}),
often on basis of its spectrum~\cite{Verkruysse2008}.
However, frequency of the typical subject's movements
(head tilts, for example) often fits within the expected HR range,
generating strong false signal~\cite{Chen2018DeepMagSS,Osman2015,Verkruysse2008}.

Poh~\etal~\cite{Poh2010} tried to resolve this issue by Independent Component Analysis (ICA),
which tries to distinguish different sources of the final signal~\cite{Macwan2018,Poh2010}.

It was demonstrated, however, that even small motion of a subject
during natural interaction with the computer
causes significant accuracy decrease
(compared to controllable no-movements case),
and in-door exercise environment makes ICA almost useless.
At the same time, similar methods improved by ML (machine learning)
techniques shows much better accuracy~\cite{Moco2016,Monkaresi2014,Nowara2018,Prakash2018}.

Our work aims at studying impact of certain architectural tricks
which could contribute to modern ML-based approaches on rPPG.

In particular, we consider classification-based estimation of HR values
by convolutional network followed by two fully connected layers.

Outputs of this network, if normalized, can be treated as relative probabilities distribution
(we have single output for the every HR value with constant step).
This ``pseudo-spectrum'' is often noisy. 
In order to suppress noise-related outliers,
some processing or filtering method 
may be applied to this distribution,
for example, smoothing. 
We replace determined processing by convolutional layers assuming to get optimal filtering
procedure during training.

\section{Related works}

Almost all of recent rPPG papers include ML elements
for pre-processing, post-processing or as a main element;
often it's neural networks
and specifically convolutional neural networks~(CNN)~\cite{Ganapathy2018Taxonomy}.

Ground truth signal in most cases retrieved by either
electrocardiography (ECG)~\cite{Chen2018, Monkaresi2014, Spetlik2018}
or contact PPG~\cite{Osman2015,Villarroel2017NoncontactVS,Yu2019RecoveringRP}.
ECG is quoted to be more reliable~\cite{Spetlik2018},
while contact PPG is quoted to be closer to rPPG signal retrieved~\cite{Yu2019RecoveringRP}, making training easier.
Reproduction of ground truth signal is often the main area of ML-based training~\cite{Chen2018}.

Some mask is often applied to select the so-called ``region of interest''~(ROI) -- area of the frame image without background pixels and with most informative fluctuations~\cite{Chen2018,Monkaresi2014,Osman2015}. In most cases, the video of the face is processed.

While modern progress in deep learning techniques is believed to provide powerful tools for the rPPG, straightforward approach is facing difficulties:

\textit{Small datasets for training.} Most of the available datasets include less than 100 subjects. Often only one type of camera is used.

To compensate a little number of subjects, researchers tend to gather a lot of videos
from every single subject, which doesn\textquotesingle t seem to resolve the problem, since even with long videos the datasets are still relatively small.
To overcome this problem, transfer learning approach is used
with original data coming from other domains~\cite{Faust2012Review, Ganapathy2018Taxonomy}
or even generated out from training on mock signals~\cite{Niu2018}. Transfer learning capability is also used in~\cite{Chen2018}
to measure quality of the proposed method.

\textit{Video compression methods}, which tend to preserve details which is visible to the naked eye,
and suppress mostly invisible (and therefore meant to be not important) details.
Another problem is variable frame-rate~\cite{Fletcher2015}, often generated by video codecs trying to keep constant bit-rate, which causes jitter on periods between frames.

It was observed that non-compressed video, while impractical, is a much better source for iPPG~\cite{Spetlik2018},
and iPPG accuracy decreases linearly as compression rate increases~\cite{McDuff_2018_CVPR_Workshops, McDuff2017Compression}.

Magnification of small motion and color changes of the skin
is used to overcome problems caused by video compression
and as a method to increase general sensitivity,~\cite{Chen2018DeepMagSS,Hurter2017Cardiolens,Wu2012EulerianMagnification}.
Another quite unusual approach is to use 1-pixel camera
which has no problems with the bandwidth and therefore needs no compression~\cite{Wang2019}.

The following methods used to handle the motion of the subject and to overcome the related problems like illumination changes:
\begin{itemize}
\item ML-based detection of peaks instead of spectrum analysis~\cite{Osman2015};
\item CNN signal post-processing to extract HR information~\cite{Spetlik2018};
\item CNN pre-processing which is expected to be stable to small movements~\cite{Spetlik2018,Tang2018};
\item skin reflection model, which is expected to help with noise
caused by observational skin color changes
caused by different view angles~\cite{Chen2018};
\item attention model -- ROI building procedure which pays special attention to moving pixels of image,
also expected to distinguish smaller movements from global rigid motion~\cite{Chen2018,Kumar2015,Yang2019};
\item spatio-temporal CNN,
which is able to extract temporal-based features out of series of 2D images
but, compared to the traditional 3D convolutions,
uses significantly less parameters for training~\cite{Yu2019RecoveringRP}.
\end{itemize}
Another sources of inspiration for this paper include:
\begin{itemize}
	\item increasing of quality of events detection
	when multi-lead ECG used instead of 1-lead~\cite{Ganapathy2018Taxonomy};
	\item motion-compensated pixel-to-pixel pulse extraction sensors used to utilize spatial-redundancy of image~\cite{Fletcher2015}.
\end{itemize}

The both researches involved multichannel registration of biosignals, which led to more accurate evaluation of spatio-temporal characteristics of signals. This could be explained by the enhancement of manifestation of the common sources of the registered signals in their cross-correlations.
\section{Experimental setup}
\label{sec:exp_setup}
This section describes the self-collected dataset containing 52 videos recorded on three cameras in different motion scenarios. The preprocessed and ground truth data are publicly available~(see Section~\ref{sec:prepro}).

Three cameras were alternately used for video recording:\\
{\bf Cam$_1$:} Logitech~C920 webcam with $1920\times1080$ (Width$\times$ Height) pixels and WMV2 video codec.\\
{\bf Cam$_2$:} Microsoft~VX800 webcam with $640\times480$ pixels and WMV3 video codec.\\
{\bf Cam$_3$:} Lenovo~B590 laptop integrated webcam with $640\times480$ pixels and WMV3 video codec.

All video sequences were recorded in RGB (24-bit depth) at 15 frames per second (fps) with 60--80 seconds duration. 
Each frame contains a person\textquotesingle s face. 

\begin{figure}[t]
	\begin{center}
		\includegraphics[width=1.0\linewidth]{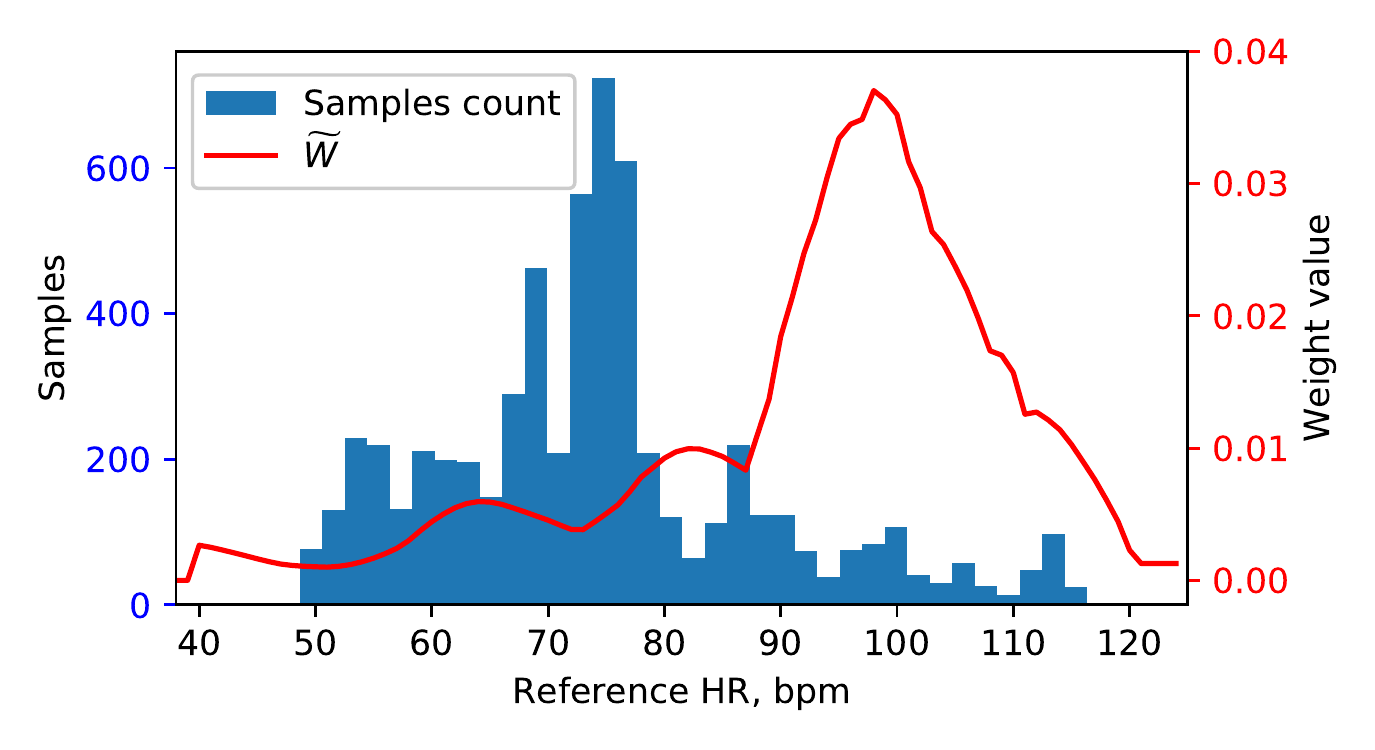}
	\end{center}
	\caption{Distribution of HR values in the experimental dataset and corresponding class weights (see Section~\ref{sec:loss}).}
	\label{fig:hr_distr}
\end{figure} 
From 2 to 14 video sequences were recorded for each of 8 healthy participants (7 male, 1 female, aged from 24 to 37, with skin-tones categorized from type-I to type-IV on the Fitzpatrick scale). Distribution of reference HR values is shown in~\figref{fig:hr_distr}. Each subject signed written consent to take part in the tests, which were performed in compliance with the bioethics regulations; experimental protocols were approved by the bioethics committee of the \ificcvfinal Southern Federal \else anonymized \fi University.

The distances range from the face to webcam was 0.5--0.7 m. The pixel size of the facial area was from $350\times350$ pixels to $550\times550$ when using the Cam$_1$ and from $150\times150$ to $250\times250$ when using cameras Cam$_2$, Cam$_3$. Each video sequence was recorded at 15 frames per second in daylight illumination (300--1000 lx).
Ground truth, or reference, HR values were obtained by the Choicemmed MD300C318 pulse oximeter (with declared mean absolute error of 2 bpm).

\begin{figure}[t]
	\begin{center}
		\includegraphics[width=1.0\linewidth]{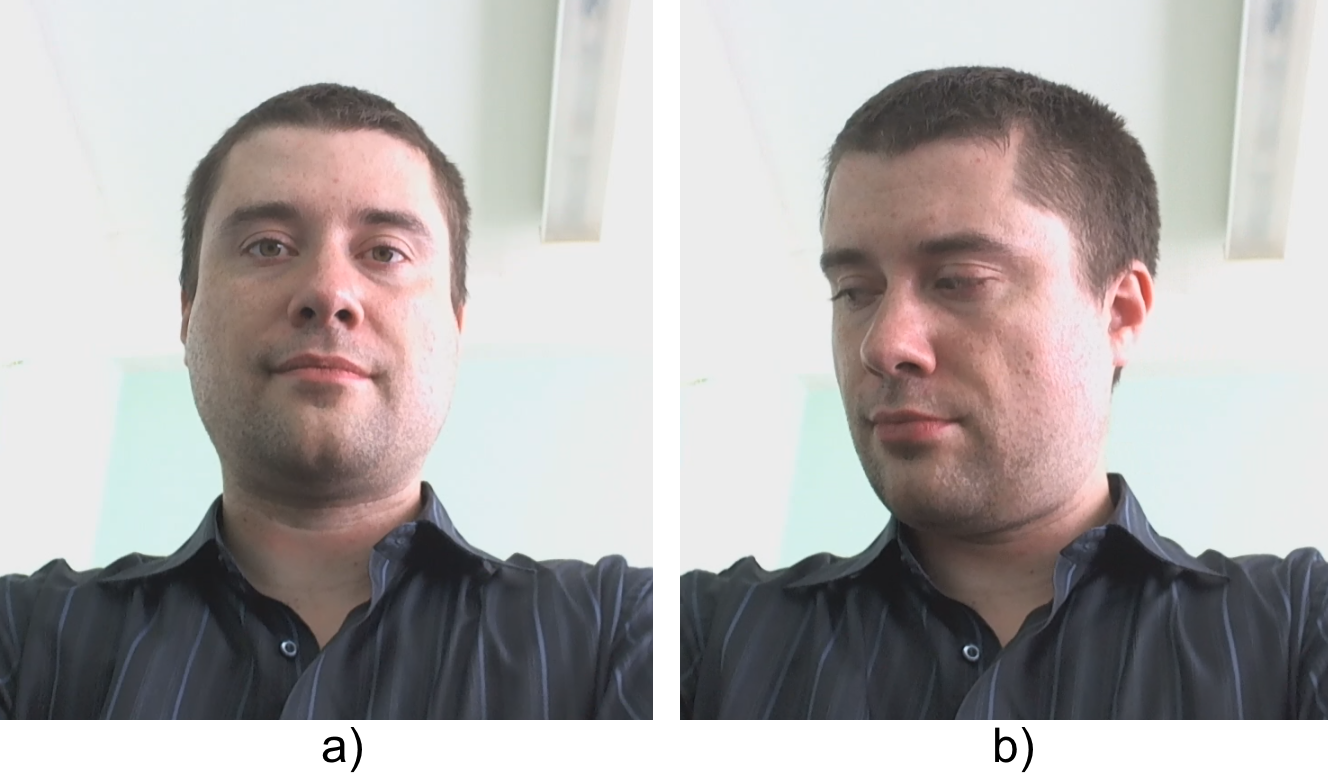}
	\end{center}
	\caption{Examples of subject\textquotesingle s poses and recording conditions in dataset: Stationary Scenario a) and Mixed Motion Scenario b).}
	\label{fig:faces}
\end{figure} 
Experiments were conducted in Stationary Scenario and in Mixed Motion Scenario~(\figref{fig:faces}).

\textbf{Stationary Scenario.} Subject sat still in front of the webcams in a fixed pose looking straight ahead. 12 video sequences for each webcam were recorded.

\textbf{Mixed Motion Scenario.}
Subject rotated their head from right to left (with $120^\circ$ amplitude), from up to down (with $100^\circ$ amplitude). Subject was asked to speak and change facial expressions. 6 video sequences were recorded for each webcam.
\section{Methods}
\begin{figure*}[t]
	\begin{center}
		\includegraphics[width=1.0\linewidth]{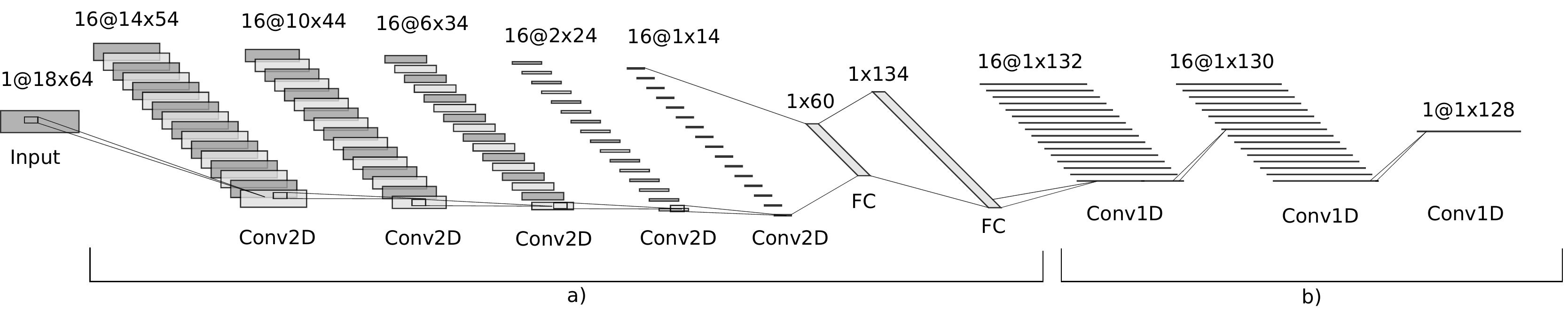}
	\end{center}
	\caption{Architecture of CL+F: basic classification network a), and filtering appendix b).}
	\label{fig:full_net}
\end{figure*}

In this section, we describe methods of data pre-processing and HR estimation on a color signal sample by means of CNN model. The final model architecture sequentially performs three steps: feature extraction (convolutional layers in~\figref{fig:full_net}~(a)\,), HR prediction (fully connected layers in~\figref{fig:full_net}~(b)\,), and filtering (\figref{fig:full_net}\,(b)\,).

The following contributions are considered.

\textbf{Multiple ROIs}, forming several input signals~~(Section~(\ref{sec:inp})\,). Our simple attention model focuses on easy-identifiable parts of the face, known to be important~\cite{Kopeliovich2016, Kumar2015, McDuff_2018_CVPR_Workshops}. CNN expected to use spatial-redundancy as in~\cite{Fletcher2015} and extract cross-correlations between signals like in multi-lead ECG~\cite{Ganapathy2018Taxonomy}.

\textbf{Pseudo-spectrum} instead of regression-like models~(Section~(\ref{sec:architecture})\,).
As long as different sources (blinking, head tilts, mimics) generates noise of different frequencies.
Therefore features important to filter them out may differs for different heart rates.
Also classification models (compared to regression ones, like used by Spetlik~\etal~\cite{Spetlik2018})
are mentioned to be effective for events detection
(HR estimation is believed to be based on detection of a heartbeat events)
even if the input is noisy~\cite{Ganapathy2018Taxonomy}. 

\textbf{Combined loss}, which is based on the cross-entropy and mean squared error losses~(Section~(\ref{sec:loss})\,). In order to minimize the penalty of the minor missclassification (when estimated HR value is close to reference one), we add a mean squared error to the loss function.

\textbf{Post-processing 1D CNN}~(Section~(\ref{sec:filt})\,).
Magnification of pseudo-spectrum peaks increases contrast
and make detection of the HR more accurate.
It resembles the post-processing approach of Spetlik~\etal~\cite{Spetlik2018}
(but used to increase contrast between classes)
and magnification techniques described in ~\cite{Chen2018DeepMagSS,Hurter2017Cardiolens}
(but used for post-processing instead of pre-processing).

\subsection{Data pre-processing}
\label{sec:prepro}
The pre-processing is made independently on each given video sequence. It includes extraction of color signals from a sequence and generating of training, validation and test sets. It is assumed that each frame of a video sequence contains face of the same person, while persons can differ in different sequences.

The data containing $\mathbf{CS}^v$ and coordinates of $\mathit{ROI}_r$ with synchronized reference HR values are publicly available~\refdataset.
\subsubsection{Color signal extraction}
\begin{figure}[t]
	\begin{center}
		\includegraphics[width=1.0\linewidth]{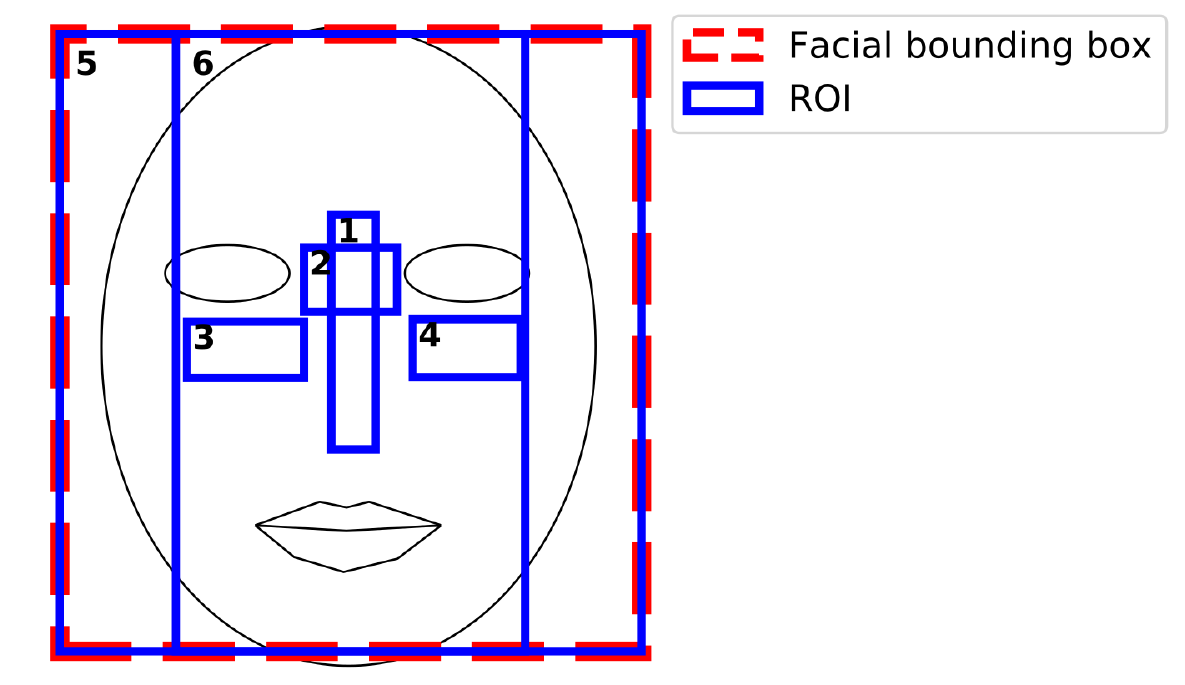}
	\end{center}
	\caption{Location of ROIs on the facial bounding box: nose (1), nose bridge (2), areas under eyes (3,\hspace{0.2em}4), truncated facial box~(5), full bounding box (6). ROIs coordinates are not ideally symmetrical about a horizontal axis: they were set manually to match the facial features on most video sequences in the dataset.}
	\label{fig:ROIs}
\end{figure} 
For a given video sequence, $\mathit{ROI}_r(t)$ is defined as $r$\nobreakdash-th rectangle with coordinates relative to facial bounding box in $t$\nobreakdash-th frame. Facial bounding box is detected by the OpenCV implementation of the Viola--Jones face detector~\cite{Viola2001} applied to each frame. Regarding the works on ROIs selection~\cite{Kopeliovich2016, Kumar2015}, six ROIs are used in this paper: $r=1..6$ (\figref{fig:ROIs}): corresponding to nose, nose bridge, areas under eyes, truncated facial box, and full bounding box. Coordinates of the bounding box are averaged over the last 20 frames ($\approx$1.3~sec) to minimize detection jitter.

$\mathit{CS}_r^c(t)$ $(c\in{R,G,B})$ is a color signal value obtained by averaging intensity of red $(R)$, green $(G)$, or blue $(B)$ color component over the $\mathit{ROI}_r(t)$. Finally, $18~(6\times 3)$ one-dimensional color signals $\mathit{CS}_r^c(t)$ were obtained from each video sequence, forming multi-dimensional color signals $\mathbf{CS}^v$, where $v$ is index of video sequence.
\subsubsection{Input data generation}
\label{sec:inp}
The input data samples are 64-frames fragments of color signals $\mathbf{CS}^v$ ($\approx$4.3~sec per fragment), obtained by splitting each signal into overlapping segments, starting from the first video frame with a step of 10 frames. Samples are scaled to fit the $\left[-1,1\right]$ interval. It is assumed that HR will not change significantly over the sample, so ground truth HR values are averaged within sample resulting in one reference value for each sample.

\begin{table}
	\begin{center}
		\footnotesize
		\begin{tabular}{|l|ccc|}
			\hline
			\multirow{2}{*}{Camera} & \multicolumn{3}{|c|}{Scenario}\\
			\cline{2-4}
			& Stationary & Mixed Motion & Both \\
			\hline\hline
			Cam$_1$ & 1169 / \enspace94 / 340 & \enspace439 / 27 / 128 & 1608 / 121 / \enspace468 \\
			Cam$_2$ & \enspace944 / \enspace64 / 279 & \enspace446 / 28 / 132 & 1390 / \enspace92 / \enspace411 \\
			Cam$_3$ & 1056 / \enspace84 / 312 & \enspace402 / 24 / 120 & 1458 / 108 / \enspace432 \\
			\hline
			All & 3169 / 242 / 931 & 1287 / 79 / 380 & \textbf{4456 / 321 / 1311}\\
			\hline
		\end{tabular}
	\end{center}
	\caption{Number of samples in the training / validation / test sets per camera and scenario.}
	\label{table:samples}
\end{table}
The video sequences are typically collected in similar environmental conditions but with different participants. In order to keep the training, validation and test sets statistically equivalent, the training set includes first 70\% of samples obtained for a $\mathbf{CS}^v$, while the validation set includes next 10\% samples (excluding ones that overlap with the training), and test set includes last 20\% samples. In this way, color signals from each video sequence are presented in all sets, while the training set doesn\textquotesingle t intersect with validation or test sets. Alternative distribution of train, validation, and test sets was also evaluated, where the sets were chosen from different non-overlapping video sequences, recorded by different webcams to estimate model generalization properties.  Table~\ref{table:samples} represents number of samples per each camera and scenario. 

The data augmentation was made on the training set, where random uniform noise was added to $\mathbf{CS}^v$ values. The noise amplitude was also a uniformly distributed random value with amplitude from 5e-3 to 5e-2, that typically corresponds to 5\%--50\% of the pulse signal amplitude. The amplitude changed after each training step.
\subsection{Network architecture}
\label{sec:architecture}
Data sample size is $(18\times64)$, where the 1st dimension is for color signal channels, the 2nd is for discrete time. A sample is processed as a single-channel image. Due to relatively large kernels and, therefore, quick reducing of temporal information through the convolutional layers of the network, we don\textquotesingle t use pooling layers to avoid double reducing.

The basic architecture contains five 2D convolution layers with ReLU activation functions followed by two fully connected layers, also with ReLU activations (see \figref{fig:full_net}(a)\,). After each convolution layer there is 2D Batch Normalization \cite{Ioffe2015}, while after fully connected layers there are 1D Batch Normalization and dropout layers (with 0.5 dropout rate). We tried to add Batch Normalization before and after ReLU, and the latter proved to lead to better accuracy. The number of output channels in convolution layers is 16. Kernel size is 5$\times$11 (color signal channels$\times$discrete time) for the first four layers, and 2$\times$11 for 5th layer. Consequently, a 16-channel image of 1$\times$14 size is input to the first fully connected layer, which has 60 output neurons.

We formulate the problem of HR estimation in two ways: as regression or classification tasks. For regression, output of the second fully connected layer is a single value $\hat{y}$ representing HR estimate. For classification, the output \out\, is a $N$-length prediction vector, where $N = 128$ is the number of classes. Classes are generated from the range of admissible HR values (40--125~bpm), which is split into $N$ segments of equal size ($\approx 0.7$~bpm). The segments are assigned to corresponding class labels \out$_i$. The resulting label $y$ is calculated as argmax$(\outf_i)$.

Note that reference value $y$ and estimate $\hat{y}$ correspond to HR values in regression task, while in classification task they are class labels. 
\subsection{Loss functions} 
\label{sec:loss}
We consider several loss functions: squared error (SE) for regression task; cross entropy (CE) and combined loss (CL) for classification task. 

SE loss is calculated between model output $\hat{y}$ and reference HR value $y$:
\begin{equation}
\text{SE}\left(\hat{y}, y\right) = (\hat{y}-y)^2
\end{equation}

The distribution of reference HR values in a dataset can be unbalanced. To compensate this, weight coefficients $\tilde{w}_y \in\widetilde{W}$ are involved in all CE losses. First, $L$ vector is calculated denoting inverse numbers of samples per classes in dataset. Next, the weights vector $\widetilde{W}$ is calculated by smoothing the $L$ vector, which is computed as:
\begin{equation}
\label{eq:smooth}
\widetilde{W} = \frac{L \ast G^k(\sigma^{2})}{\sum_{i=1}^{128}L \ast G^k(\sigma^{2})},
\end{equation}
where $G^k$ is zero-mean Gaussian kernel with window size $k=13$~bpm and $\sigma=\frac{13}{3}\approx 4.3$~bpm. Operator $\ast$ means discrete convolution with the same padding.

The CE loss combines softmax and negative log likelihood functions:
\begin{equation}
\begin{aligned}
\text{CE}\left(\outf, y\right) &= -\tilde{w}_y \ln\left( \frac{\exp\outf_y}{\sum_{i=1}^{128}{\exp{\outf_i}}} \right) = \\
							   &= \tilde{w}_y\left( -\outf_y + \ln\left(
											 						\sum_{i=1}^{128}{\exp{\outf_i}}\right) 
								      \right)
\end{aligned}
\end{equation}

In pure classification task, loss value does not depend on distance between predicted and reference HR values. To take account for the distance, we introduce one-hot vector $\text{Y}^y \left( \in \{0,1\}^{\mathbb{N}}: Y^y_i = 1 \Leftrightarrow i=y \right)$, corresponding to $y$ label. Similarly to \eqreff{eq:smooth}, the vector is smoothed resulting in $\widetilde{\text{Y}}^y$:
\begin{equation}
\label{eq:smooth2}
\widetilde{\text{Y}}_y = \frac{\text{Y}^y \ast G^k(\sigma^{2})}{\sum_{i=1}^{128}\text{Y}^y \ast G^k(\sigma^{2})},
\end{equation}
where $k=13$~bpm, $\sigma=\frac{k}{6}\approx 2.2$~bpm.

Finally, combined loss CL is calculated based on CE loss:
\begin{equation}
\label{eq:CL}
\text{CL}\left(\outf, y\right) = \text{CE}\left(\outf, y\right) +
						   \alpha \cdot \text{MSE}\left( \outf, \widetilde{\text{Y}}^y\right),
\end{equation}
where $\text{MSE}(\outf, \widetilde{\text{Y}}^y) = \frac{1}{N}\sum_{i=1}^N(\outf_i - \widetilde{\text{Y}}^y_i)^2$ is a mean squared error, $\alpha$ is a balancing coefficient equal to $25$. The coefficient is selected heuristically in order to equalize the contribution of terms to the sum in \eqreff{eq:CL} during training.
\subsection{Filtering}
\label{sec:filt}
CL loss function implies a comparison between model prediction \out\, and smoothed one-hot vector $\widetilde{\text{Y}}^y$ (see \eqreff{eq:CL}\,). Smoothing parameters $\sigma$ and $k$ (\eqreff{eq:smooth2}\,) can be hyperparameters, however, their optimization is a challenging problem that should be possibly done after each training iteration. Instead of adding new hyperparameters, we add filtering step to the basic network architecture (\figref{fig:full_net}(b)\,), optimizing smoothing of \out\, during standard training process in order to fit $\widetilde{\text{Y}}^y$.

Filtering step goes after the output of second fully connected layer. We use three 1D convolution layers with 16 output channels (single channel for the last layer) and kernel size of $3$. Each layer is followed by ReLU activation and 1D Batch Normalization. Due to the layers containing no padding, output of the second fully connected layer increased to $134$, so the model output shape remains the same single-channel vector of $N$ length.

We recommend using the filtering step together with CL~loss. Nevertheless, it also can be applied when using CE~loss in order to clarify a class label. Regarding the regression task, where the model output is a single value, the filtering step is inappropriate.
\section{Experimental evaluation and results}
\label{sec:res}
This section describes metrics for models evaluation, list of training hyperparameters, and experimental results.
\subsection{Evaluation metric}
We used two metrics to evaluate the performance of different models of HR estimation. All metrics were applied to results on color signal samples of a test set.
\paragraph{MAE}
Mean absolute error calculates L1-distance between estimated vector $\widehat{\textbf{Y}}$ of HR values and referent vector~$\textbf{Y}$:
\begin{equation}
\label{eq:mae}
\text{MAE}(\widehat{\textbf{Y}}, \textbf{Y}) = \frac{1}{M}\sum_{j=1}^M\left|\widehat{\textbf{Y}}_j - \textbf{Y}_j\right|,
\end{equation}
where M is a number of samples. We treat MAE as a qualitative measure of model accuracy.
\paragraph{Coverage at $\pm$3 bpm}
This metric was used by Wang~\etal~\cite{Wang2019} for analysis of video sequences. We redefine it as a percentage of samples for which MAE value was smaller than 3 bpm. Regarding the classification task, model output is one of $128$ classes corresponding to segments within the range of admissible HR values (40--125~bpm). Therefore, for the classification, we use coverage at $\pm$4 class labels which are approximately equal to $\pm$2.7~bpm. Coverage metric can be interpreted as model quality. The 3~bpm threshold is close to the MAE of the pulse oximeter (2~bpm) indicating that such a threshold could be used to determine an acceptable measurement.
\subsection{Hyperparameters}
Here we describe hyperparameters used during the training process. In this work, we set them manually and don\textquotesingle t address their optimization.

The batch size was 1024 samples; the number of epochs was 5000. The training set was randomly shuffled after each epoch. The best model parameters were selected from epoch with minimum MAE value on the validation set. The optimization method was Adam~\cite{KinBa17} with default parameters. The balancing coefficient $\alpha$ for the CL metric (Eq.~\eqref{eq:CL}\,) is also a hyperparameter.
\\[-8mm]
\paragraph{Learning rate} 
\begin{figure}[t]
	\begin{center}
		\includegraphics[width=0.9\linewidth]{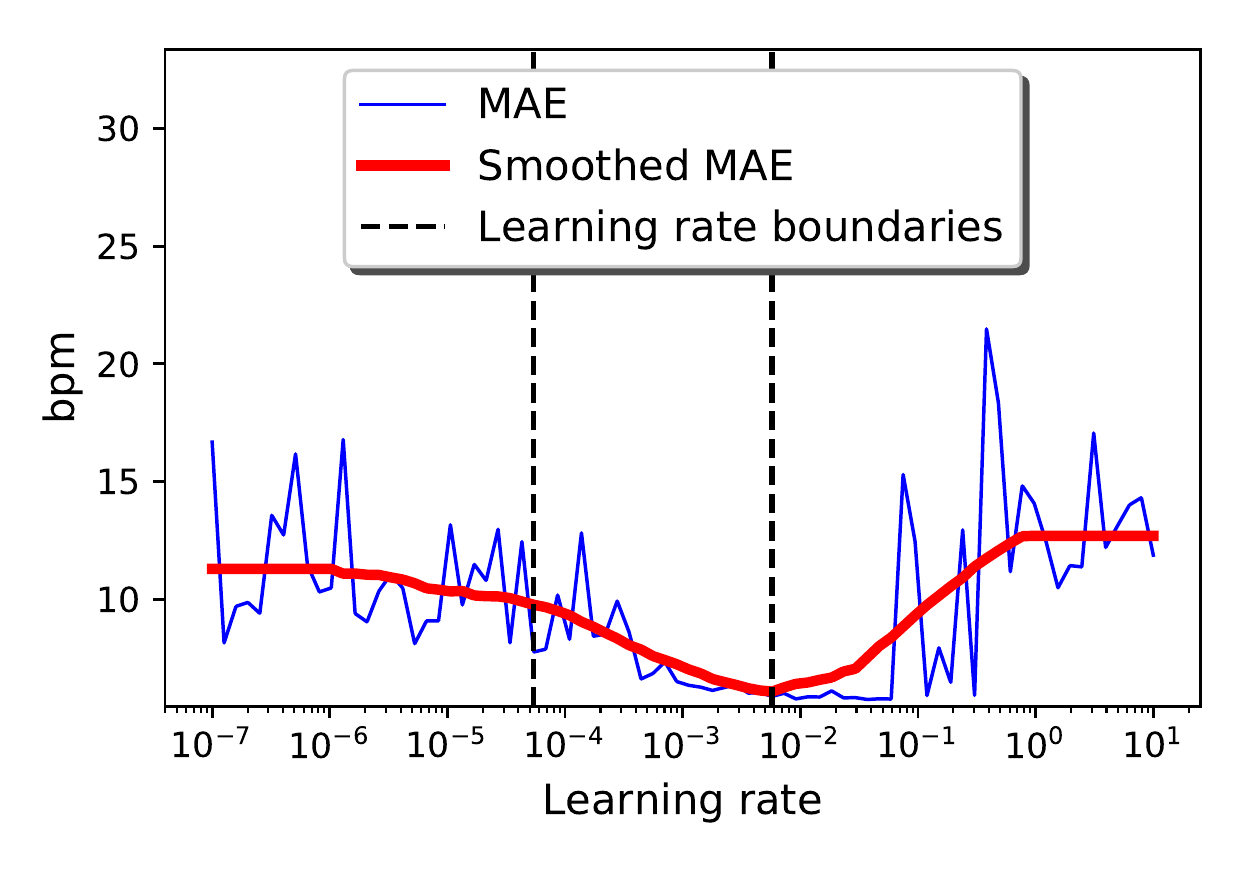}
	\end{center}
	\caption{Classification loss as a function of learning rate (for CL+F). The dashed lines indicate the minimum and maximum learning rates (5.8e-5 and 5.8e-3 respectively).}
	\label{fig:lr}
\end{figure} 
Before the training process, we applied the learning rate range test~\cite{Smith2015} to choose the learning rate boundaries. The test consisted in the estimation of MAE metric after 5-epochs training for several learning rates varying from 10e-7 to 10e+1. The resulting curve (see \figref{fig:lr}) was smoothed using Gaussian kernel. The maximum learning rate is defined as argmin of the smoothed curve; minimum learning rate is chosen by dividing the maximum reduced by two orders of magnitude. During further training, learning rate was linearly changed from minimum to maximum and back according to the ``1cycle'' learning policy~\cite{Smith2017}.
\subsection{Results}
We evaluated four models that are titled by the corresponding loss functions:
SE model for regression task (62,675~parameters);
CE (70,676~parameters),
CL~(70,676~parameters),
and CL+F (with filtering layers, 72,017~parameters) for the classification task.

Training and evaluation methods were implemented in Python (using PyTorch library).
The code for generating of the dataset from the color signals
as well as the implementation of the proposed architecture, training and testing procedures,
and trained models are freely available online~\refdataset.

\begin{table*}
	\begin{center}
		\small
		\begin{tabular}{|l|cc|cc|cc|cc|cc|cc|}
			\hline
			\textit{Test subset} & \multicolumn{2}{|c|}{Stationary} & \multicolumn{2}{|c|}{Mixed Motion} & \multicolumn{2}{|c|}{Cam$_1$} & \multicolumn{2}{|c|}{Cam$_2$} & \multicolumn{2}{|c|}{Cam$_3$} & \multicolumn{2}{|c|}{Full test set}\\
			\cline{1-13}
			\textit{Model} & MAE & Cover & MAE & Cover & MAE & Cover & MAE & Cover & MAE & Cover & MAE & Cover\\
			\hline \hline
			SE 			 & 11.6 & 10.8\% & 15.4 & 13.7\% & 13.3 & 13.7\% & 11.7 & 12.4\% & 12.9 & 9.0\% & 12.7 & 11.7\% \\
			CE 			 & 4.4 & 44.8\% & 11.1 & 31.3\% & 5.8 & 36.8\% & 7.5 & 38.7\% & 5.6 & 47.2\% & 6.3 & 40.9\% \\
			CL 	 & \textbf{3.8} & \textbf{48.3\%} & 8.9 & 39.5\% & 5.5 & 43.8\% & 6.4 & 34.5\% & \textbf{3.9} & \textbf{58.1\%} & 5.3 & 45.8\% \\
			CL+F & 4.1 & 47.8\% & \textbf{6.9} & \textbf{48.7\%} & \textbf{4.6} & \textbf{44.7\%} & \textbf{5.6} & \textbf{45.0\%} & 4.5 & 54.2\% & \textbf{4.9} & \textbf{48.1\%} \\
			\hline
		\end{tabular}
	\end{center}
	\caption{Evaluation results of the considered models. The models were trained on the full training set and evaluated on the subsets of the test set. Cover means Coverage at $\pm$3~bpm metric. The best values are in bold.}
	\label{table:res_main}
\end{table*}
The test set was divided into the several subsets by scenario and used cameras
(see Section~\ref{sec:exp_setup}):
Stationary and Mixed Motion subsets titled according to the corresponding scenarios,
Cam$_1$, Cam$_2$, Cam$_3$ subsets containing the samples from the corresponding cameras in both scenarios,
and Full test set, which includes all samples of the test set.
\subsubsection{Accuracy on test subsets}
Model comparison results are presented in Table~\ref{table:res_main}.
The considered models were trained on the Full training set (defined in Section \ref{sec:inp}).
Then the models were evaluated on the test subsets.
It is clear that the SE model had much lower accuracy than classification-based models.
Accuracy of both CL and CL+F models was higher than of CE, where the distance between classes is not taken into account.
Adding filtering layers to the the CL model led to the highest accuracy in most cases including Full test set.
The CL model had low coverage value on Mixed Motion and Cam$_2$ subsets.
The former can be explained by the presence of high-amplitude noise in color signals caused by motions.

The coverage metric estimations of the CL+F model were typically near 50\%.
It is insufficient to use the model in a practical applications.
However, the size of the training set was nearly 15 times smaller than the number of the model parameters.
Therefore, the model accuracy and coverage could grow with the dataset expansion.

\begin{figure*}[t]
	\begin{center}
		\includegraphics[width=1.0\linewidth, height=0.65\linewidth]{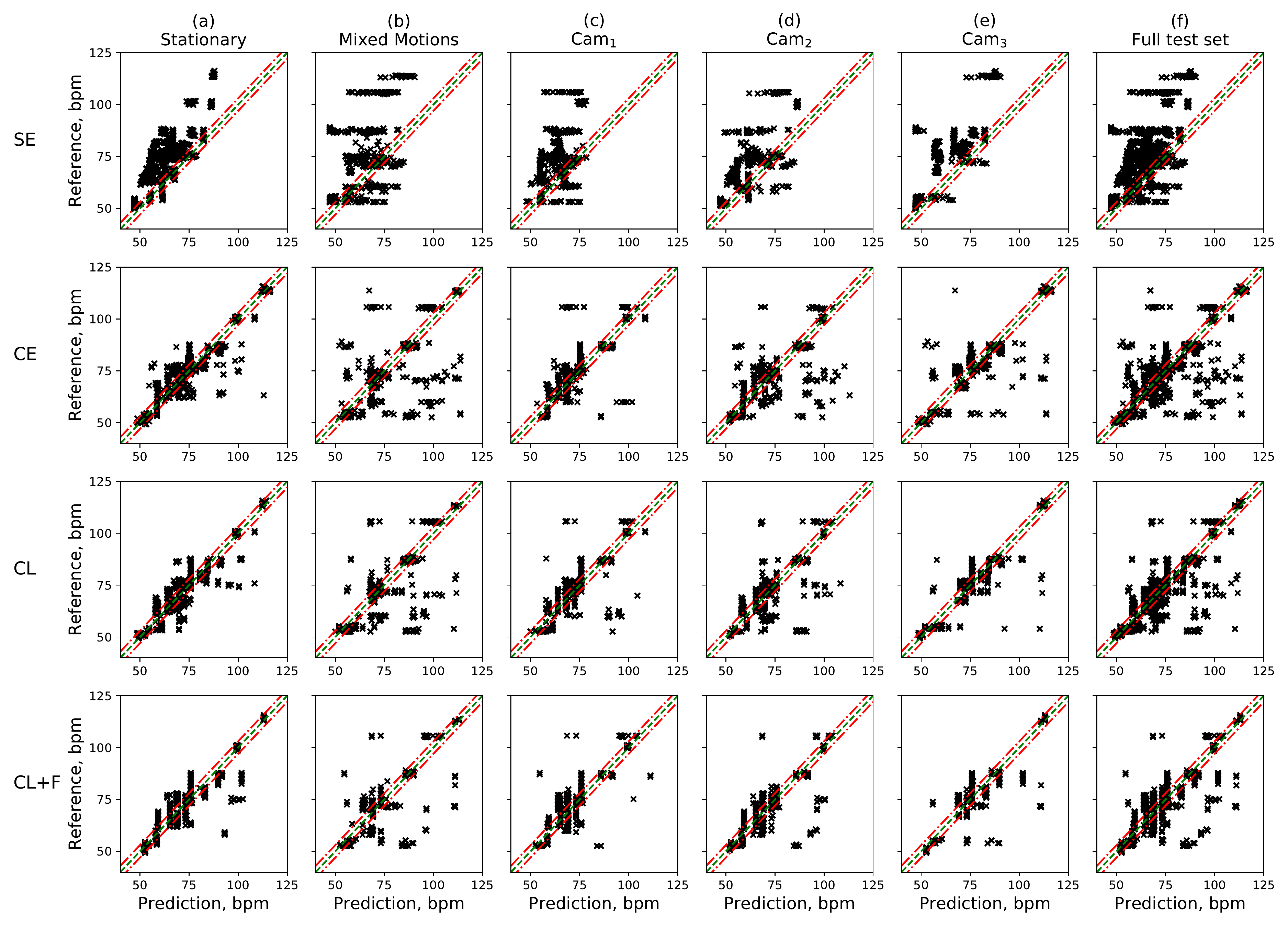}
	\end{center}
	\caption{Scatter plots of reference HR values and HR estimations of models trained on the full training set. From top to bottom: models SE, CE, CL, CL+F. From left to right: test subsets a) Stationary, b) Mixed Motion, c) Cam$_1$ only, d) Cam$_2$ only, e) Cam$_3$ only, f) Full test set. Dashed line is a zero-error line, two dash-dotted lines form error interval of 3~bpm.}
	\label{fig:scatter}
\end{figure*} 
\figref{fig:scatter} shows scatter plots for the considered models evaluated on the different test subsets. Predictions of the SE model were distributed within the first half of admissible HR range as the result of the unbalanced dataset (see \figref{fig:hr_distr}). Classification-based models led to the similar plots differing in a number of outliers, where the CL+F model showed the best results.

As said before, head motion causes disturbances in the color signal. The amplitude of disturbances is up to two orders higher than amplitude of the signal. Due to that, it had been unexpected for the model to have the same coverage values on the Stationary and Mixed Motion subsets, which was true for CL+F. Different MAE values indicate a large number of outliers (\figref{fig:scatter}(b)\,) on the Mixed Motion subset. We believe that the filtering out of such outliers merits further research.
\subsubsection{Model generalization}
\begin{table}
	\begin{center}
		\small
		\begin{tabular}{|l|c|c|c|c|c|}
			\hline	
			\multirow{2}{*}{Model} & \multicolumn{4}{|c|}{Test subset}\\
			\cline{2-5}
			& Cam$_1$ & Cam$_2$ & Cam$_3$ & Full test set\\
			\hline \hline
			(CL+F)$_1$ & 4.9 & 13.3 & 17.2 & 11.6\\
			(CL+F)$_2$ & 20.5 & 5.0 & 16.6 & 14.3\\
			(CL+F)$_3$ & 18.2 & 20.3 & 4.8 & 14.4\\
			\hline
			(CL+F)$_{1,2}$ & 3.1 & 4.6 & 14.8 & 7.5\\
			(CL+F)$_{1,3}$ & 7.6 & 13.5 & 7.1 & 9.3\\
			(CL+F)$_{2,3}$ & 10.1 & 5.5 & 4.6 & 6.8\\
			\hline
		\end{tabular}
	\end{center}
	\caption{MAE values of CL+F trained on different training subsets.}
	\label{table:res_pairs}
\end{table}
We studied the generalization of the CL+F network architecture by comparing of the models trained on the different subsets: (CL+F)$_i$ ($i\in\left\{1,2,3\right\}$) trained on the training subsets with samples from the single camera Cam$_i$; (CL+F)$_{i,j}$ ($i,j\in\left\{1,2,3\right\}, i\ne j$) trained on the training subsets with the samples from two cameras Cam$_i$, Cam$_j$. Sets based on two cameras were reduced by removing random 50\% of the samples in order to equalize a number of samples.

MAE values are given in the Table~\ref{table:res_pairs}. In the single-camera case, the error was high on the every test subset excluding one corresponding to the camera. This is due to the different camera resolutions, noise, codecs, and other parameters. Two-cameras case led to the similar results: low error for cameras from the training subset and high errors for the remaining camera. However, the error for the remaining camera was noticeably lower than the errors on cameras out of training subset in the single-camera case. Moreover, two-camera cases provided better accuracy on the full test set. As the every training subset had a comparable number of samples, we conclude that the CL+F network architecture provides a high generalizing ability for its instances.
\section{Conclusion}
The problem of remote photoplethysmography by means
of deep learning was considered. Color signals, which are
time series of red, green, and blue color components averaged
over certain regions in the facial area (cheeks, forehead,
nose, \etc.), were used as inputs. Inputs were processed by convolutional
neural network followed by two fully connected
layers. Multiple outputs of this network correspond to different
possible HR values, with constant step. The impact
of improvements to network architecture and loss function
was studied.

In particular, adding convolutional-based filter
for post-processing of network outputs led to better accuracy of HR estimations. 
We expect this improvement can benefit to wide range of 
deep neural network architectures which address a regression problem by classification 
and produce ``pseudo-spectrum'' as output.

Another improvement is the combined loss function, where the first component is a cross entropy
and the second one is a mean squared error between the network output
and smoothed one-hot vector. The proposed model demonstrated 
generalization tendency: the model performance, which was 
evaluated on a particular camera increases with 
an increasing number of cameras in the training set (excluding 
the chosen camera); the number of training samples preserved the same.
\ificcvfinal
\\[-8mm]
\paragraph{Acknowledgments.} The project is supported by the Russian Ministry for Education and Science, project no.2.955.2017/4.6 ``Development of the hardware and software system for monitoring the attention level and psychoemotional state of pilots and dispatching personnel to improve flight safety''.
\fi

{\small
\bibliographystyle{ieee}
\bibliography{egpaper_final}
}

\end{document}